\documentclass{article}

\usepackage[numbers]{natbib}

    \usepackage[preprint]{neurips_2024}

\usepackage[utf8]{inputenc} 
\usepackage[T1]{fontenc}    
\usepackage{hyperref}       
\usepackage{url}            
\usepackage{booktabs}       
\usepackage{amsfonts}       
\usepackage{nicefrac}       
\usepackage{microtype}      
\usepackage{xcolor}         

\usepackage{adjustbox}
\usepackage{multirow}
\usepackage{colortbl}
\usepackage{amsmath}
\usepackage{tabularx}
\usepackage{subcaption}

\usepackage{graphicx}
\usepackage{graphics}   
\usepackage{amssymb}
\usepackage{balance}
\usepackage{adjustbox}
\usepackage{algorithm}
\usepackage{algorithmic}
\usepackage{tikzducks}
\usepackage{pifont}

\newcommand{\Fig}[1]{Figure~\ref{fig:#1}}

\def\ie{\textit{i.e.}}
\def\eg{\textit{e.g.}}

\definecolor{brown}{rgb}{0.85, 0.15, 0.15}
\definecolor{purp}{rgb}{0.95, 0.16, 0.65}
\definecolor{purpc}{rgb}{0.95, 0.36, 0.65}
\definecolor{darkpurple}{rgb}{0.54, 0.17, 0.89}
\definecolor{orange}{rgb}{0.9, 0.45, 0.0}
\definecolor{blue}{rgb}{0.0, 0.5, 1.0}
\definecolor{green}{rgb}{0, 0.8, 0}
\definecolor{darkgreen}{rgb}{0, 0.6, 0}
\definecolor{lgreen}{rgb}{0.6, 0.8, 0}
\definecolor{red}{rgb}{0.8, 0, 0}
\definecolor{redd}{rgb}{0.9, 0, 0}
\definecolor{yellow}{rgb}{0.75, 0.56, 0}
\definecolor{darkblue}{rgb}{0.2, 0.2, 0.8}
\definecolor{brinkpink}{rgb}{0.98, 0.38, 0.5}
\definecolor{cadmiumred}{rgb}{0.89, 0.0, 0.13}
\definecolor{ceruleanblue}{rgb}{0.16, 0.32, 0.75}
\definecolor{dandelion}{rgb}{0.94, 0.88, 0.19}
\definecolor{bostonuniversityred}{rgb}{0.8, 0.0, 0.0}
\definecolor{brown(web)}{rgb}{0.65, 0.16, 0.16}
\definecolor{cornellred}{rgb}{0.7, 0.11, 0.11}
\definecolor{greend}{rgb}{0.0, 0.35, 0.0}

\definecolor{lblue}{rgb}{0, 0.2, 0.8}
\definecolor{dorange}{rgb}{0.8, 0.4, 0.0}

\def\etal{\textit{et al.}}

\definecolor{err}{rgb}{0.99, 0, 0}

\definecolor{grey}{rgb}{0.9, 0.9, 0.9}

\newcommand{\hyperfootnote}[1][]{\def\ArgI\hyperfootnoteRelay}
\newcommand\hyperfootnoteRelay[2][]{\href{#1#2}{\ArgI}\footnote{\href{#1#2}{#2}}}

\usepackage{enumitem}

\title{GaRA-SAM: Robustifying Segment Anything Model with Gated-Rank Adaptation}

\author{%
  Sohyun Lee$^{1}$ \hspace{3mm} Yeho Gwon$^{1}$ \hspace{3mm} Lukas Hoyer$^{2}$ \hspace{3mm} Suha Kwak$^{1}$ \vspace{3mm}\\
    $^1$POSTECH \qquad \ \       
    $^2$Google\\
  }

\begin{document}

\maketitle

\begin{abstract}
Improving robustness of the Segment Anything Model (SAM) to input degradations is critical for its deployment in high-stakes applications such as autonomous driving and robotics. Our approach to this challenge prioritizes three key aspects: first, parameter efficiency to maintain the inherent generalization capability of SAM; second, fine-grained and input-aware robustification to precisely address the input corruption; and third, adherence to standard training protocols for ease of training. To this end, we propose gated-rank adaptation (GaRA). 
GaRA introduces lightweight adapters into intermediate layers of the frozen SAM, where each adapter dynamically adjusts the effective rank of its weight matrix based on the input by selectively activating (rank-1) components of the matrix using a learned gating module. This adjustment enables fine-grained and input-aware robustification without compromising the generalization capability of SAM. 
Our model, GaRA-SAM, significantly outperforms prior work on all robust segmentation benchmarks.  
In particular, it surpasses the previous best IoU score by up to 21.3\%p on ACDC, a challenging real corrupted image dataset.

\end{abstract}

\section{Introduction}

\begin{figure}[t!]
    \centering
    \includegraphics[width=\linewidth]{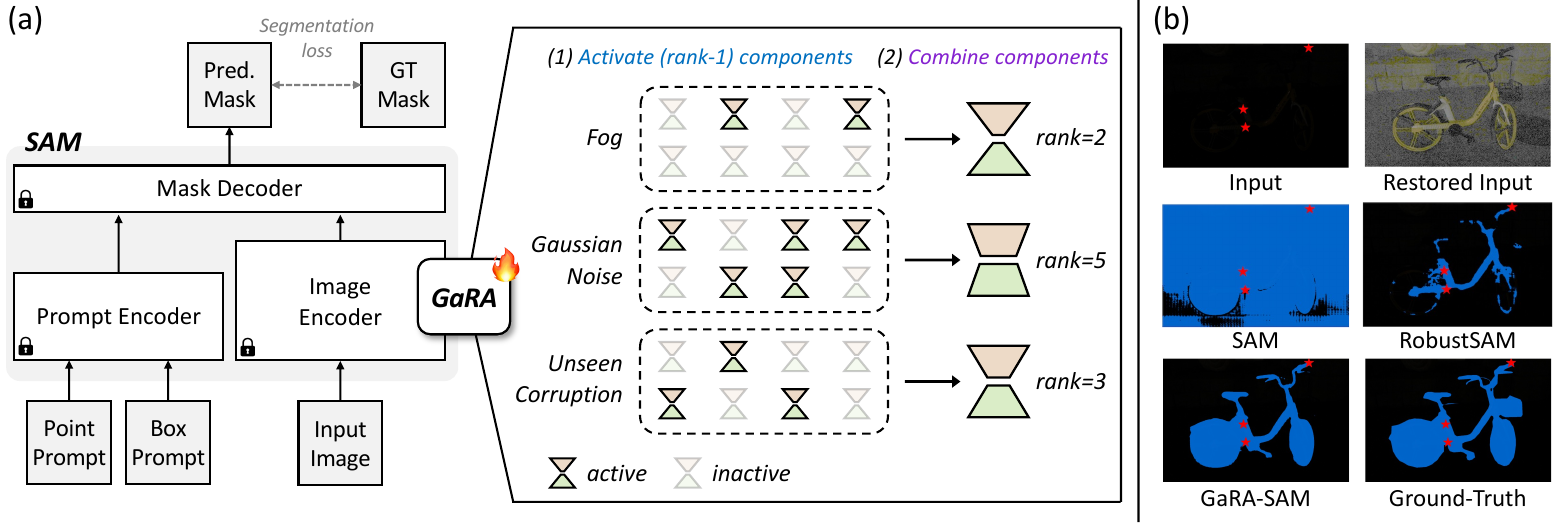}
    \vspace{-5mm}
    \caption{Overview and example results of GaRA-SAM.
    (a) Conceptual illustration of GaRA-SAM. (b) Example results on a real low-light image~\cite{lis}. GaRA-SAM produces an accurate mask while the original SAM fails. 
    Note that the restored input is provided for illustrative purposes only; GaRA-SAM does not perform image restoration. 
    }
    \vspace{-2mm}
    \label{fig:teaser}
\end{figure}

The Segment Anything Model (SAM)~\cite{kirillov2023segment} has proven to be a powerful tool for zero-shot image segmentation, exhibiting impressive generalization across unseen objects and images without additional training. Nevertheless, its performance deteriorates considerably when faced with degraded input due to noise, blur, low illumination, and adverse weather~\cite{chen2024robustsam}, as shown in \Fig{teaser}(b). This limitation significantly restricts its use in high-stakes applications such as autonomous driving and robotics.

A seemingly straightforward approach to improving the robustness of SAM is to attach an existing image restoration module to the front of SAM. However, this typically introduces significant computational overhead, and often yields suboptimal segmentation performance since image restoration is optimized to enhance the perceptual quality of images, rather than to improve performance of segmentation models like SAM~\cite{chen2022rvsl,diamond2021dirty,lee2022fifo,li2019single,son2020urie,vidal2018ug}. 
An alternative strategy involves fine-tuning SAM entirely on degraded inputs, which, however, demands substantial computational resources and diminishes its inherent zero-shot generalization capability~\cite{chen2024robustsam}.

Chen~\etal~\cite{chen2024robustsam} addressed these limitations by introducing anti-degradation modules into SAM;
these modules are trained to refine degraded features by promoting feature-level consistency between a pair of clean and corrupted images of the same content. 
Despite effectively enhancing robustness, their method, dubbed RobustSAM, has a couple of limitations. First, its requirement for paired clean and degraded images, which cannot be captured by typical cameras, necessitates the use of synthetic degradations in training, hindering its generalization to real-world degradations.
Second, since RobustSAM aims at learning representations invariant to various degradations, it struggles to produce representations adapted to the specific degradation affecting the input at hand, restricting further performance improvement.
These limitations reduce the effectiveness of RobustSAM in real-world scenarios, where inputs can be degraded by real and previously unseen corruptions. 

We found that low-rank adaptation (LoRA)~\cite{hu2022lora} offers a promising foundation for overcoming many of the previously discussed limitations. 
By introducing and training lightweight adapters within the frozen SAM, LoRA improves the robustness parameter-efficiently while preserving the generalization ability of SAM. 
Our empirical analysis confirms that SAM incorporating LoRA achieves impressive robustness without using paired clean and degraded images for training, thus establishing a strong baseline as is.
Nevertheless, our analysis also reveals a key limitation: the fixed rank of the adapters hinders its effective adaptation to a wide range of corruptions.  
We found a significant variation in the optimal rank of the adapters between different corruption types and inputs, underscoring the need for input-adaptive rank modulation.

Based on this observation, we introduce \emph{gated-rank adaptation} (GaRA), which is illustrated in \Fig{teaser}(a), a novel method designed to enhance the robustness of SAM while addressing the aforementioned limitations. GaRA dynamically adjusts the effective rank of the weight matrix of each adapter, while being parameter-efficient and not demanding clean-degraded image pairs for training.
Specifically, GaRA decomposes the weight matrix of an adapter into (rank-1) components and chooses a proper subset of them dynamically according to the input. 
To achieve this, we introduce a gating module that predicts a binary vector selectively activating the most appropriate components for the input.
This mechanism allows GaRA to flexibly control both the number and combination of active components based on the input
without any test-time optimization. 
Consequently, our zero-shot segmentation model integrating GaRA with SAM, named GaRA-SAM, achieves fine-grained and input-aware robustification while remaining parameter-efficient.

GaRA-SAM achieves the state of the art on multiple robust segmentation benchmarks~\cite{lis,ChengPAMI,lvis,coco,acdc,street,ndd,bdd}, including both synthetic and real-world corruption datasets. 
Importantly, and in contrast to prior work, the design of GaRA-SAM enables training using real-world degraded images lacking clean counterparts, leading to notable performance improvement on real-world corruption benchmarks.
The main contribution of this work is three-fold: 
\begin{itemize}[leftmargin=5mm] 
    \setlength\itemsep{1mm}
    \item Our extensive analysis reveals the surprising effectiveness of LoRA in robustifying SAM, with its optimal rank varying significantly across different corruption types and individual images. These findings suggest a new research avenue for improving robustness of vision foundation models.
    \item We propose GaRA, a novel method to achieve robust SAM. At its core lies a lightweight and input-dependent adapter that enables fine-grained and parameter-efficient robustification without compromising the generalization capability of SAM. Also, GaRA does not require paired clean and degraded images for training and thus, unlike previous work~\cite{chen2024robustsam}, it can be learned using real degraded images without their clean references.
    \item Our final model, GaRA-SAM, achieves the state of the art across multiple robust segmentation benchmarks. Notably, it significantly surpasses the previous best IoU score by up to 21.3\%p on ACDC, a challenging real-world corrupted image dataset. 
\end{itemize}

\section{Related Work}

\noindent\textbf{Robust Segment Anything.}
SAM~\cite{kirillov2023segment} accepts free-form prompts along with an image to produce relevant masks, showing superior zero-shot generalizability. 
Despite its success, its robustness against visual corruptions is questionable~\cite{chen2024robustsam,schiappa2024robustness}. To improve the robustness of SAM, RobustSAM~\cite{chen2024robustsam} adopts anti-degradation modules to approximate features of clean images. However, it requires clean-degraded image pairs for training, which are often unavailable in real-world scenarios. 
Improving the robustness of vision models with image restoration~\cite{ai2024lora,AirNet,luo2024controlling,potlapalli2023promptir,son2020urie} and degradation-specific techniques~\cite{Bi_2024_CVPR,lis,lee2022fifo,ma2022both} has also been investigated. 
AirNet~\cite{AirNet} and URIE~\cite{son2020urie} are universal image restoration models, but they introduce heavy computational overhead and AirNet targets a better image quality, not improving performance of visual perception models. 
LoRA-IR~\cite{ai2024lora}, DA-CLIP~\cite{luo2024controlling}, and PromptIR~\cite{potlapalli2023promptir} tackle this by encoding degradation-specific information into prompts. However, they rely on additional information such as degradation types and textual descriptions, and suffer from their complex modules and training schemes. 
Meanwhile, FIFO~\cite{lee2022fifo} and FreD~\cite{Bi_2024_CVPR} focus on extracting fog-invariant features, making their models robust to only foggy scenes. 
In contrast to these prior arts, GaRA is an efficient yet effective approach to fine-tuning SAM for improving its robustness without demanding clean references of degraded images for training.

\noindent\textbf{Low-rank Adaptation.}
Fine-tuning large-scale pre-trained models introduces intensive
overheads in space and time. 
To overcome this, various parameter-efficient learning schemes such as prompt tuning~\cite{jia2022vpt,lester2021power,li2021prefix,shen2024multitask} and adapter-based fine-tuning~\cite{adaptformer,qlora,he2021towards,houlsby2019parameter,hu2021lora,kim2024efficient,scalingshifting,pfeiffer2020adapterfusion,yin20245100breakingperformanceshackles} have emerged.
Among these, LoRA~\cite{hu2021lora} leverages 
trainable low-rank matrices to catch up with full fine-tuning while introducing marginal extra learnable parameters. 
However, fixed ranks of LoRA yield limitations such as poor generalization in certain tasks~\cite{valipour-etal-2023-dylora,xia2024chain} and inefficient parameter allocation~\cite{chang2025elalora,zhang2023adaptive}. Despite the efforts of previous work to update the ranks during training dynamically~\cite{lin2024nora,valipour-etal-2023-dylora,zhang2023adaptive}, they still use fixed ranks at test time and require handcrafted rank selection. 
In contrast, GaRA learns a small module that selects and combines appropriate (rank-1) components of a LoRA block, adapting to individual samples with various visual corruptions even during inference.

\noindent\textbf{Mixture of Experts.}
While model scaling has been known as an effective and promising way of constructing a powerful model, training such a model on large-scale datasets~\cite{alabdulmohsin2022revisiting,he2016deep,kaplan2020scaling,kolesnikov2020big,pmlr-v97-tan19a,zhai2022scaling} is challenging due to high computational requirements~\cite{fedus2022switch,shazeer2017}. 
To this end, mixture of experts (MoE), which adopts multiple submodules and considers each as an expert, has gained prominence~\cite{fedus2022switch,MLSYS2023_5616d34c,lin2024moe,shazeer2017,wang-etal-2022-adamix,zhou2022mixture}. These submodules are active or inactive using a learnable gating module at both training and test time, resulting in efficient use of resources and improved training stability. 
Switch Transformer~\cite{fedus2022switch} sparsely simplifies the MoE paradigm and proposes to select experts sparsely.
AdaMix~\cite{wang-etal-2022-adamix} injects an MoE adapter consisting of various up- and down-sampling layers into each transformer layer, both improving parameter-efficiency and performance. 
Motivated by the prior work, GaRA first decomposes the low-rank matrices of LoRA into (rank-1) components and considers each as an expert. Then, a learnable gating module sparsely selects a combination of (rank-1) components in an input-adaptive manner, enabling adaptation on a per-input basis.

\section{Proposed Method}

We first describe LoRA as a strong baseline for robustifying SAM in Sec.~\ref{sec:lora}, and analyze the impact of the rank of its adapters on the robustness of SAM to verify our motivation of input-adaptive rank modulation in Sec.~\ref{sec:lora_analysis}. 
Finally, Sec.~\ref{sec:gara} details GaRA that we integrate into SAM as GaRA-SAM.

\subsection{Foundation: Low-rank Adaptation for Robustifying SAM} \label{sec:lora}

LoRA~\cite{hu2022lora} is a parameter-efficient adaptation method that imbues a frozen pretrained model with adaptability using a handful of trainable parameters.
It freezes a pretrained weight matrix $\mathbf{W}_0 \in \mathbb{R}^{D \times K}$ and introduces trainable low-rank update $\Delta \mathbf{W} = \mathbf{B}\mathbf{A}$, where $\mathbf{B} \in \mathbb{R}^{D \times R}$ and $\mathbf{A} \in \mathbb{R}^{R \times K}$ with rank $R \ll \min(D, K)$.
The adapted forward pass becomes:
\begin{equation}
    \mathbf{h} = \mathbf{W}_0 \mathbf{x} + \mathbf{B}\mathbf{A}\mathbf{x},
\end{equation}
where $\mathbf{x} \in \mathbb{R}^K$ is the input and $\mathbf{h} \in \mathbb{R}^D$ is the output.

\begin{figure}[t!]
    \centering
    \includegraphics[width=1.0\linewidth]{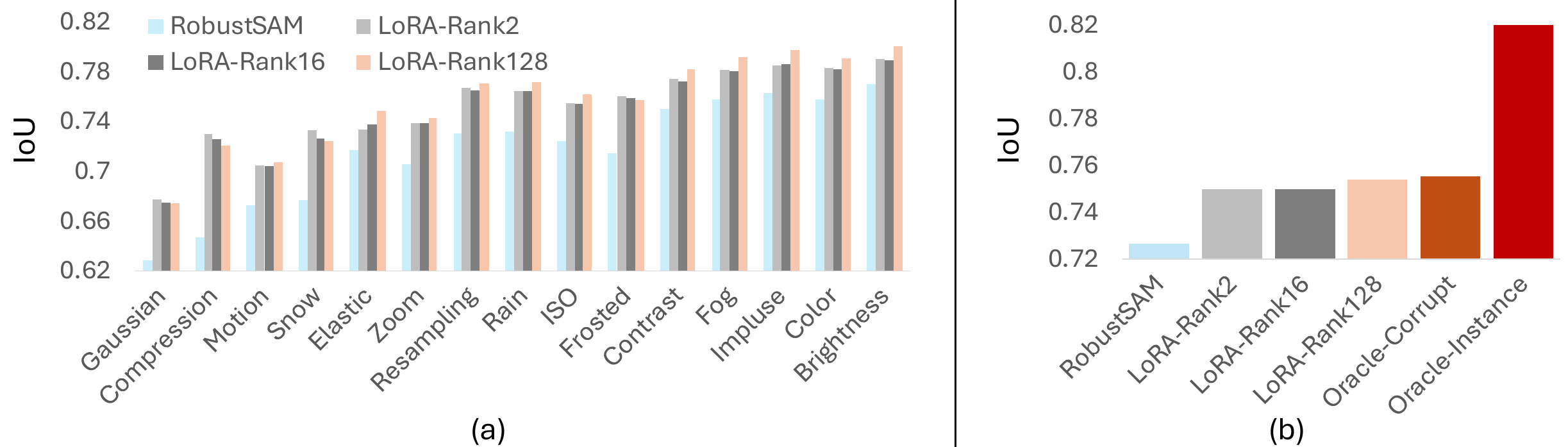}
    \vspace{-5mm}
    \caption{Impact of the rank in LoRA integrated with SAM. The models were evaluated on LVIS~\cite{lvis} using point prompts.
    (a) Performance versus rank under various corruption types. 
    The best rank varies depending on the corruption type.
    (b) Comparisons of rank selection strategies. Oracle-Corrupt chooses the best rank per corruption type, while Oracle-Instance selects the best rank per image. The outstanding performance of Oracle-Instance suggests the need for input-adaptive rank manipulation.
    }
    \label{fig:lora_rank}
    \vspace{-2mm}
\end{figure}

Given its original purpose of parameter-efficient adaptation without compromising generalization, we argue that LoRA offers a reasonable baseline for improving robustness of SAM against input degradation. 
To validate this potential empirically, we evaluated SAM integrated with LoRA for robust segmentation using point prompts following an evaluation protocol of RobustSAM~\cite{chen2024robustsam}. 
Specifically, we froze the original weights of SAM, attached low-rank adapters $\Delta \mathbf{W}$ to the key, query, and value projection layers of its image encoder, and trained the adapters with the standard segmentation loss on degraded images.
As demonstrated in \Fig{lora_rank}, SAM with LoRA clearly outperformed RobustSAM~\cite{chen2024robustsam} on the LVIS dataset~\cite{lvis}, although it does not require clean reference images and the auxiliary loss the RobustSAM demands.

\subsection{Analysis on the Impact of Rank on Robustness}\label{sec:lora_analysis}
Since different degradations distort different aspects of the input, the level of representation capacity needed for robustifying SAM varies accordingly, motivating the investigation into the role of LoRA's rank. To analyze this,
we first evaluated its performance across various input corruptions while varying the rank. 
The results in \Fig{lora_rank}(a) suggest that the optimal rank varies depending on the corruption type.
We conjecture that more pronounced corruptions lead to more substantial contamination of the semantic content of the input, and lower-rank adapters are more effective in such conditions since their restricted capacity forces them to prioritize more essential features for segmentation~\cite{liu2023vida}.

To further verify this conclusion, we evaluated SAM with LoRA under two \emph{oracle} rank modulation scenarios: selecting the best rank per corruption type (Oracle-Corrupt) and per image (Oracle-Instance) using ground-truth. 
Specifically, we first computed IoU scores for all candidate ranks and selected only the top-performing rank for each corruption type or each image. 
While Oracle-Corrupt yielded moderate gains over fixed ranks, Oracle-Instance achieved significantly higher performance, highlighting the substantial potential of fine-grained, input-adaptive rank selection.
These findings motivate our design of GaRA, which dynamically adjusts the rank per input.

\subsection{Gated-rank Adaptation}
\label{sec:gara}

\begin{figure}[t!]
    \centering
    \includegraphics[width=\linewidth]{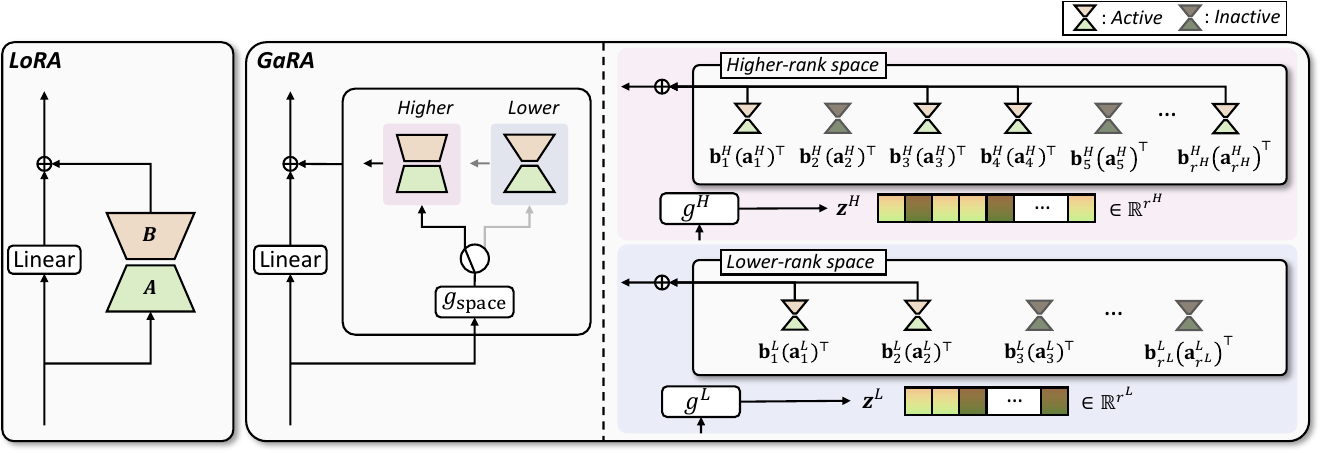}
    \vspace{-5mm}
    \caption{
    Adapter architecture of GaRA and comparison to LoRA. GaRA leverages hierarchical gating for both coarse and fine control over the adaptation process.
    First, $g_\text{space}$ selects between the higher-rank and lower-rank spaces based on the input. Then, the corresponding gating module $g^{H}$ or $g^{L}$ predicts a binary vector $\mathbf{z}^{H}$ or $\mathbf{z}^{L}$ to activate a subset of (rank-1) components tailored to the input. These active components are composed to form the final update matrix for adaptation.}
    \label{fig:architecture}
    \vspace{-2mm}
\end{figure}

Our solution for robustifying SAM, termed gated-rank adaptation (GaRA), is designed with three key properties in mind: parameter efficiency to preserve the generalization ability of SAM, fine-grained and input-aware robustification, and adherence to standard segmentation learning protocols for ease of training.
These properties are realized by a novel, lightweight adapter that dynamically adjusts itself based on the input corruption; 
the architecture of the adapter is depicted in \Fig{architecture}.
Our final model, GaRA-SAM, is constructed by integrating these adapters into the key, query, and value projection layers of the image encoder of SAM, keeping the original SAM weights frozen.
Also, GaRA-SAM is trained solely with the standard segmentation loss computed from degraded images, without auxiliary learning objectives or paired clean references.

GaRA introduces a gating strategy to modulate the adapter's rank based on the input. 
To this end, we reinterprete the update matrix of LoRA, \ie, $\Delta \mathbf{W} = \mathbf{B}\mathbf{A}$,
as a composition of $R$ (rank-1) components: $\Delta \mathbf{W} = \sum_{i=1}^{R} \mathbf{b}_i \mathbf{a}_i^\top,$
where $\mathbf{b}_i \in \mathbb{R}^{D}$ and $\mathbf{a}_i \in \mathbb{R}^{K}$ are the $i$-th column and row of $\mathbf{B}$ and $\mathbf{A}$, respectively. 
We assume that 
$R$ is 
large enough
so that the model has access to a rich set of (rank-1) components, enabling
GaRA to flexibly choose a relevant subset depending on the input.

GaRA employs a hierarchical gating strategy that provides both coarse and fine control over the adaptation process. The adapter first coarsely selects between lower- and higher-rank spaces based on the input, and then, within the selected rank space, the gating module finely determines the rank of the adapter by activating a subset of the (rank-1) components appropriate for the input. 
The coarse rank space selection separates the adapter's components into two exclusive sets, tailored to different degraded inputs demanding different representation capacities, to mitigate potential conflicts between the sets and further improve performance.
Meanwhile, the fine-grained gating enables flexible and input-specific composition of the update matrix, supporting efficient and expressive adaptation to diverse corruptions. Below we elaborate on this gating process.

\noindent\textbf{Gating 1: Rank Space Selection.}
We explicitly divide the adapter into two distinct rank spaces, a lower-rank set $\{ \mathbf{a}_i^L, \mathbf{b}_i^L \}_{i=1}^{r_L}$ and a higher-rank set $\{ \mathbf{a}_j^H, \mathbf{b}_j^H \}_{j=1}^{r_H}$ with $r_L < r_H \ll K$, where $K$ denotes the input feature dimension, and $r_L$ and $r_H$ indicate the maximal ranks of the lower-rank and higher-rank spaces, respectively.\footnote{We use the terms `lower-rank' and `higher-rank' in a relative sense. Both $r_L$ and $r_H$ are substantially smaller than the full feature dimension $K$, and thus still lie in the low-rank regime.}
To choose between these rank spaces, we employ a binary gating module $g_{\text{space}}$, which takes as input the intermediate feature $f(\mathbf{x})$ computed from $\mathbf{x}$ and outputs a binary gating variable $z_{\text{space}} \in \{0,1\}$, where 0 and 1 indicate the lower- and higher-rank space, respectively.
This module consists of a two-layer MLP followed by the Gumbel-Sigmoid, allowing differentiable binary decisions during training.
The gating process is formally expressed as:
\begin{equation}
\alpha_\text{space} = \text{MLP}_{\text{space}}(f(\mathbf{x})),
\end{equation}
where $\alpha_\text{space}$ is the gating logit computed from the input feature.
To enable backpropagation despite the binary nature of the binary gating variable $z_{\text{space}}$, we apply the Gumbel-Sigmoid relaxation~\cite{jang2016categorical}: 
\begin{equation}
\tilde{z}_\text{space} = \sigma\left(\frac{1}{\tau} (\alpha_\text{space} + G)\right),
\end{equation}
where $G \sim \text{Gumbel}(0,1)$ is a noise sampled during training, $\sigma$ is the sigmoid function, and $\tau$ is a temperature parameter controlling the sharpness of the sigmoid.
During training, we apply hard thresholding in the forward pass as $z_\text{space} = \mathbb{I}[\tilde{z}_\text{space} > 0.5]$ while using the continuous value $\tilde{z}_\text{space}$ to compute gradients in the backward pass.
At test time, we discard the noise and compute the gate: $z_\text{space} = \mathbb{I}[\sigma(\alpha_\text{space}) > 0.5]$.

\noindent\textbf{Gating 2: (rank-1) Component Selection.}
Within the selected rank space, 
the associated gating module predicts a binary vector identifying (rank-1) components suitable for the input:
\begin{equation}
\mathbf{z}^L = g^L(f(\mathbf{x})) \in \{0,1\}^{r_L}, \quad \mathbf{z}^H = g^H(f(\mathbf{x})) \in \{0,1\}^{r_H}.
\end{equation}
where $g^L(\cdot)$ and $g^H(\cdot)$ denote the gating modules for the lower-rank and higher-rank spaces, respectively.
Each gating module consists of a three-layer MLP whose final output is a real-valued logit vector, followed by Gumbel-Sigmoid operations to obtain binary masks:
\begin{equation}
\boldsymbol{\alpha}^L = \text{MLP}_L(f(\mathbf{x})), \quad \boldsymbol{\alpha}^H = \text{MLP}_H(f(\mathbf{x})).
\end{equation}
To enable gradient-based learning, we apply the Gumbel-Sigmoid relaxation to each element of the logit vectors~\cite{jang2016categorical}:
\begin{equation}
\tilde{z}_i = \sigma\left(\frac{1}{\tau} (\alpha_i + G_i)\right), \quad i = 1, \dots, r_L \text{ or } r_H,
\end{equation}
where $\alpha_i$ is the $i$-th logit from $\boldsymbol{\alpha}^L$ or $\boldsymbol{\alpha}^H$, $G_i \sim \text{Gumbel}(0,1)$ is the corresponding noise sample, $\tau$ is a temperature parameter, and $\sigma$ is the sigmoid function. 
During training, hard thresholding is applied in the forward pass to produce binary decisions ($z_i = \mathbb{I}[\tilde{z}_i > 0.5]$), while the continuous $\tilde{z}_i$ is used for backpropagation. 
At inference time, the gating becomes deterministic: $z_i = \mathbb{I}[\sigma(\alpha_i) > 0.5]$.
We apply this relaxation to the gating functions $g^{L}$ and $g^{H}$ to enable learnable and input-adaptive binary decisions for the selection of (rank-1) components.
This gating mechanism results in a dynamic, input-dependent adapter update:
\begin{equation}
\Delta \mathbf{W} =
(1 - z_{\text{space}}) \cdot \sum_{i=1}^{r_L} z_i^L \mathbf{b}_i^L (\mathbf{a}_i^L)^\top +
z_{\text{space}} \cdot \sum_{j=1}^{r_H} z_j^H \mathbf{b}_j^H (\mathbf{a}_j^H)^\top.
\end{equation}
This design of GaRA enables flexible and conflict-free adaptation to a wide range of corruptions by coarse-to-fine rank modulation.

\section{Experiments}
\label{sec:exp}

\subsection{Experimental Setting}
\label{sec:exp-settings}

\noindent\textbf{Dataset.}
For training and validation, we utilize the Robust-Seg dataset~\cite{chen2024robustsam}, which is constructed by applying 15 types of synthetic corruptions to three semantic segmentation benchmarks: LVIS~\cite{lvis}, MSRA-10K~\cite{ChengPAMI}, and ThinObjects-5K~\cite{liew2021deep}, comprising a total of 26,000 masks.
For evaluation, we use five clear-condition image segmentation benchmarks: LVIS, MSRA-10K, STREETS~\cite{street}, NDD20~\cite{ndd}, COCO~\cite{coco}. Also, we test on a real-world corrupted benchmark including BDD-100K~\cite{bdd} and LIS~\cite{lis}.
For training GaRA-SAM on real-world data, we use BDD-100K and LIS for training, and evaluate on BDD-100K and LIS, and ACDC~\cite{acdc}.
Note that all these datasets are free from licensing issues.
For brevity, we refer to the union of BDD-100K and LIS as BDD+LIS, and the union of STREETS and NDD20 as STREETS+NDD.
More details are given in the appendix.
\noindent\textbf{Experimental Details.}
We adopt the ViT-B and ViT-L variants of SAM~\cite{dosovitskiy2020image}, and freeze their parameters during training the GaRA modules.
The models are optimized by Adam~\cite{kingma2014adam} with a learning rate of $1 \times 10^{-4}$ for ViT-B and $1 \times 10^{-5}$ for ViT-L, a weight decay of $1 \times 10^{-5}$, and input batches of size 8, using both point and box prompts. 
The gating modules are trained with the same learning rate. 
We set the lower- and higher-rank dimensions as $r_L = 16$ and $r_H = 256$, respectively, and use a Gumbel-Sigmoid temperature of 0.5.
Since no official evaluation code is provided by previous work, we reproduce and evaluate its models using the same protocol. 
All training and evaluation experiments were conducted at POSTECH.

\subsection{Comparison on Seen Dataset}
To assess the robustness of GaRA-SAM, we first evaluate its performance on synthetic corruptions applied to seen datasets such as LVIS and MSRA-10K. The results in Table~\ref{tab:lvis-msra} show that GaRA-SAM outperforms prior methods, including RobustSAM~\cite{chen2024robustsam}, HQ-SAM~\cite{sam_hq}, and restoration-based methods (\eg, AirNet~\cite{AirNet} and URIE~\cite{son2020urie}), when using both point and box prompts in both datasets. Notably, our model demonstrates significant improvements in the degraded setting, \eg, surpassing the previous best by up to 4.3\%p on LVIS without compromising performance on clean images.

\begin{table}[t!]
    \centering
    \huge
    \renewcommand{\arraystretch}{1.1}
    \caption{Segmentation results on LVIS and MSRA using point and box prompts.}
    \resizebox{\textwidth}{!}{
    \begin{tabular}{ll|cccccccc|cccccccc}
        \toprule
        \multirow{3}{*}{Backbone} & \multirow{3}{*}{Method} 
        & \multicolumn{8}{c|}{\textbf{LVIS}} 
        & \multicolumn{8}{c}{\textbf{MSRA}} \\
        \cmidrule(lr){3-10} \cmidrule(lr){11-18}
        & & \multicolumn{4}{c}{Point Prompts} & \multicolumn{4}{c|}{Box Prompts} 
          & \multicolumn{4}{c}{Point Prompts} & \multicolumn{4}{c}{Box Prompts} \\
        \cmidrule(lr){3-6} \cmidrule(lr){7-10} \cmidrule(lr){11-14} \cmidrule(lr){15-18}
        & & \multicolumn{2}{c}{Degrade} & \multicolumn{2}{c}{Clear} 
          & \multicolumn{2}{c}{Degrade} & \multicolumn{2}{c|}{Clear} & \multicolumn{2}{c}{Degrade} & \multicolumn{2}{c}{Clear} 
          & \multicolumn{2}{c}{Degrade} & \multicolumn{2}{c}{Clear} \\
        \cmidrule(lr){3-4} \cmidrule(lr){5-6} \cmidrule(lr){7-8} \cmidrule(lr){9-10} \cmidrule(lr){11-12} \cmidrule(lr){13-14} \cmidrule(lr){15-16} \cmidrule(lr){17-18}
        & & IoU & Dice & IoU & Dice
          & IoU & Dice & IoU & Dice
          & IoU & Dice & IoU & Dice
          & IoU & Dice & IoU & Dice \\
        \midrule

        \multirow{6}{*}{\textit{ViT-B}} 
        & SAM & 65.0 & 76.3 & 70.1 & 80.1 & 75.5 & 84.5 & 78.4 & 86.1 & 75.9 & 84.5 & 79.1 & 86.6 & 86.6 & 92.2 & 88.7 & 93.4 \\
        & HQ-SAM & 69.5 & 80.1 & 76.0 & 84.8 & 79.0 & 87.1 & 83.1 & 89.7 & 82.8 & 89.6 & 86.6 & 92.0 & \underline{89.7} & \underline{94.3} & \underline{92.4} & \underline{95.9} \\
        & AirNet+SAM & 64.8 & 76.1 & 70.1 & 80.1 & 75.4 & 84.4 & 78.3 & 86.0 & 75.7 & 84.3 & 79.1 & 86.6 & 86.4 & 92.1 & 88.8 & 93.4 \\
        & URIE+SAM & 64.8 & 76.2 & 70.0 & 80.0 & 74.5 & 83.7 & 78.0 & 85.8 & 74.6 & 83.5 & 77.6 & 85.6 & 85.9 & 91.8 & 88.8 & 93.5 \\
        & RobustSAM & \underline{72.7} & \underline{82.6} & \underline{77.2} & \underline{85.8} & \underline{81.5} & \underline{89.0} & \underline{84.3} & \underline{90.7} & \underline{86.6} & \underline{92.3} & \underline{89.6} & \underline{94.2} & 89.5 & 94.2 & 92.1 & 95.7 \\
        & \textbf{GaRA-SAM} & \textbf{77.0} & \textbf{85.7} & \textbf{81.3} & \textbf{88.7} & \textbf{83.7} & \textbf{90.5} & \textbf{86.1} & \textbf{91.9} & \textbf{89.4} & \textbf{94.0} & \textbf{91.4} & \textbf{95.2} & \textbf{92.3} & \textbf{95.8} & \textbf{93.9} & \textbf{96.8} \\

        \midrule

        \multirow{6}{*}{\textit{ViT-L}} 
        & SAM & 66.2 & 76.0 & 75.0 & 83.0 & 79.9 & 87.7 & 82.8 & 89.4 & 77.3 & 84.6 & 82.0 & 88.1 & 87.6 & 92.9 & 88.9 & 93.5 \\
        & HQ-SAM & 72.6 & 82.1 & 79.0 & 86.7 & 81.1 & 88.6 & 84.7 & 90.8 & 85.0 & 91.0 & 87.6 & 92.6 & 89.4 & 94.1 & 91.4 & 95.2 \\
        & AirNet+SAM & 66.0 & 75.7 & 74.8 & 82.8 & 79.7 & 87.6 & 82.7 & 89.4 & 76.9 & 84.3 & 81.7 & 87.8 & 87.5 & 92.8 & 89.0 & 93.6 \\
        & URIE+SAM & 66.4 & 76.3 & 74.8 & 83.0 & 79.4 & 87.4 & 82.7 & 89.4 & 78.1 & 85.4 & 83.0 & 88.9 & 87.6 & 92.8 & 89.6 & 94.0 \\
        & RobustSAM & \underline{75.6} & \underline{84.5} & \underline{80.0} & \underline{87.5} & \underline{83.6} & \underline{90.3} & \underline{85.8} & \underline{91.6} & \underline{87.6} & \underline{92.9} & \underline{90.1} & \underline{94.4} & \underline{91.6} & \underline{95.4} & \underline{93.6} & \underline{96.5} \\
        & \textbf{GaRA-SAM} & \textbf{78.5} & \textbf{86.6} & \textbf{82.6} & \textbf{89.4} & \textbf{84.4} & \textbf{90.8} & \textbf{86.7} & \textbf{92.2} & \textbf{89.6} & \textbf{94.1} & \textbf{91.6} & \textbf{95.3} & \textbf{92.6} & \textbf{96.0} & \textbf{94.0} & \textbf{96.8} \\

        \bottomrule
    \end{tabular}
    }
    \label{tab:coco-sn}
    \vspace{-3mm}
\end{table}

\begin{table}[t!]
    \vspace{-3mm}
    \centering
    \huge
    \renewcommand{\arraystretch}{1.1}
    \caption{Zero-shot segmentation results on COCO and STREETS+NDD using point and box prompts.}
    \resizebox{\textwidth}{!}{
    \begin{tabular}{ll|cccccccc|cccccccc}
        \toprule
        \multirow{3}{*}{Backbone} & \multirow{3}{*}{Method} 
        & \multicolumn{8}{c|}{\textbf{COCO}} 
        & \multicolumn{8}{c}{\textbf{STREETS+NDD}} \\
        \cmidrule(lr){3-10} \cmidrule(lr){11-18}
        & & \multicolumn{4}{c}{Point Prompts} & \multicolumn{4}{c|}{Box Prompts} 
          & \multicolumn{4}{c}{Point Prompts} & \multicolumn{4}{c}{Box Prompts} \\
        \cmidrule(lr){3-6} \cmidrule(lr){7-10} \cmidrule(lr){11-14} \cmidrule(lr){15-18}
        & & \multicolumn{2}{c}{Degrade} & \multicolumn{2}{c}{Clear} 
          & \multicolumn{2}{c}{Degrade} & \multicolumn{2}{c|}{Clear} & \multicolumn{2}{c}{Degrade} & \multicolumn{2}{c}{Clear} 
          & \multicolumn{2}{c}{Degrade} & \multicolumn{2}{c}{Clear} \\
        \cmidrule(lr){3-4} \cmidrule(lr){5-6} \cmidrule(lr){7-8} \cmidrule(lr){9-10} \cmidrule(lr){11-12} \cmidrule(lr){13-14} \cmidrule(lr){15-16} \cmidrule(lr){17-18}
        & & IoU & Dice & IoU & Dice
          & IoU & Dice & IoU & Dice
          & IoU & Dice & IoU & Dice
          & IoU & Dice & IoU & Dice \\
        \midrule

        \multirow{6}{*}{\textit{ViT-B}}
         & SAM                & 65.2 & 76.3 & 69.6 & 79.6 & 76.0 & 84.8 & 78.9 & 86.4 & 74.4 & 83.5 & 81.8 & 89.1 & 80.2 & 88.3 & 85.4 & 91.8 \\
         & HQ-SAM             & 70.2 & 80.6 & 76.0 & 84.8 & 79.6 & 87.5 & 83.5 & 90.1 & \underline{75.4} & \underline{84.5} & \underline{82.2} & \underline{89.5} & 80.8 & 88.7 & 86.2 & 92.3 \\
         & AirNet+SAM         & 65.0 & 76.1 & 69.5 & 79.5 & 75.9 & 84.7 & 78.8 & 86.3 & 74.3 & 83.4 & 81.8 & 89.1 & 80.1 & 88.2 & 85.4 & 91.8 \\
         & URIE+SAM           & 65.0 & 76.2 & 69.6 & 79.6 & 75.0 & 84.0 & 78.2 & 85.8 & 74.1 & 83.4 & 81.0 & 88.6 & 79.8 & 88.0 & 84.9 & 91.5 \\
         & RobustSAM          & \underline{72.7} & \underline{82.6} & \underline{77.1} & \underline{85.7} & \underline{81.6} & \underline{89.1} & \underline{84.5} & \underline{90.9} & 74.5 & 84.1 & 81.0 & 88.8 & \underline{81.8} & \underline{89.4} & \underline{86.2} & \underline{92.4} \\
         & \textbf{GaRA-SAM} & \textbf{77.2} & \textbf{85.8} & \textbf{81.0} & \textbf{88.4} & \textbf{84.2} & \textbf{90.8} & \textbf{86.4} & \textbf{92.1} & \textbf{77.6} & \textbf{86.4} & \textbf{82.9} & \textbf{90.1} & \textbf{84.1} & \textbf{91.0} & \textbf{87.7} & \textbf{93.3} \\
         
        \midrule
        
        \multirow{6}{*}{\textit{ViT-L}}
        & SAM & 66.9 & 76.5 & 74.6 & 82.5 & 80.4 & 88.0 & 82.9 & 89.4 & 72.9 & 81.3 & 82.1 & 89.0 & 81.5 & 89.2 & 86.4 & 92.5 \\
        & HQ-SAM & 73.2 & 82.5 & 79.0 & 86.7 & 81.7 & 89.0 & 85.1 & 91.1 & \underline{76.8} & \underline{85.4} & \underline{83.9} & \underline{90.6} & 81.7 & 89.3 & 86.7 & 92.6 \\
        & AirNet+SAM & 66.6 & 76.3 & 74.4 & 82.4 & 80.3 & 87.9 & 82.8 & 89.3 & 72.5 & 81.0 & 81.9 & 88.7 & 81.4 & 89.1 & 86.4 & 92.5 \\
        & URIE+SAM & 66.6 & 76.4 & 74.3 & 82.4 & 79.9 & 87.7 & 82.8 & 89.4 & 72.6 & 81.4 & 81.2 & 88.4 & 81.1 & 88.9 & 86.1 & 92.2 \\
        & RobustSAM & \underline{75.3} & \underline{84.3} & \underline{79.9} & \underline{87.6} & \underline{83.8} & \underline{90.5} & \underline{86.2} & \underline{91.9} & 75.7 & 84.8 & 82.7 & 89.9 & \underline{83.0} & \underline{90.2} & \underline{87.5} & \underline{93.1} \\
        & \textbf{GaRA-SAM} & \textbf{78.9} & \textbf{86.9} & \textbf{82.3} & \textbf{89.1} & \textbf{84.9} & \textbf{91.2} & \textbf{86.9} & \textbf{92.4} & \textbf{80.0} & \textbf{88.0} & \textbf{85.3} & \textbf{91.7} & \textbf{85.3} & \textbf{91.7} & \textbf{88.7} & \textbf{93.9} \\

        \bottomrule
    \end{tabular}
    \vspace{-3mm}
    }
    \label{tab:lvis-msra}
\end{table}

\subsection{Zero-shot Segmentation Comparison}
To further assess the generalization capability of GaRA-SAM, we evaluate it under both synthetic and real corruptions on datasets not seen during training. Specifically, we include COCO and STREETS+NDD with synthetic degradations, and BDD+LIS, which exhibits real corruptions such as fog, rain, motion blur, and low-light.
As reported in Table~\ref{tab:coco-sn}, GaRA-SAM consistently achieves the best across all prompt types and metrics in both COCO and STREETS+NDD.
Table~\ref{tab:bdd-lis} shows that it also outperforms previous work on real corrupted images, improving over RobustSAM by more than 3.4\%p IoU with point prompts in ViT-L, despite being trained only on unpaired corrupted images.
It demonstrates that GaRA-SAM generalizes well to both synthetic and real-world corruptions, highlighting the effectiveness of its input-adaptive design.

\begin{table*}[t]
\resizebox{\linewidth}{!}{
\begin{minipage}[t]{0.45\linewidth}
\centering
\captionsetup{font=footnotesize}
\caption{Evaluation results on a real degraded image dataset, BDD+LIS, using point and box prompts.}
\vspace{-2mm}
\renewcommand{\arraystretch}{1.21}
\resizebox{1.0\linewidth}{!}{
\large
\begin{tabular}{llcccc}
        \toprule
        \multirow{2}{*}{Backbone} & \multirow{2}{*}{Method} 
        & \multicolumn{2}{c}{\textbf{Point Prompts}} 
        & \multicolumn{2}{c}{\textbf{Box Prompts}} \\
        \cmidrule(lr){3-4} \cmidrule(lr){5-6}
        & & IoU & Dice & IoU & Dice \\
        \midrule

        \multirow{6}{*}{\textit{ViT-B}} 
        & SAM               & 65.2 & 75.4 & 74.2 & 82.7 \\
        & HQ-SAM            & 67.6 & 77.9 & 69.4 & 78.0 \\
        & AirNet+SAM        & 64.1 & 74.5 & 73.4 & 82.1 \\
        & URIE+SAM          & 65.7 & 76.1 & 74.1 & 82.8 \\
        & RobustSAM         & \underline{69.5} & \underline{79.5} & \underline{75.7} & \underline{84.1} \\
        & \textbf{GaRA-SAM} & \textbf{71.3} & \textbf{80.9} & \textbf{76.8} & \textbf{85.3} \\
        
        \midrule
        
        \multirow{6}{*}{\textit{ViT-L}} 
        & SAM               & 68.9 & 77.8 & 74.3 & 82.0 \\
        & HQ-SAM            & \underline{72.7} & \underline{81.7} & 76.3 & 84.3 \\
        & AirNet+SAM        & 67.2 & 76.4 & 73.1 & 81.3 \\
        & URIE+SAM          & 69.7 & 78.8 & 74.4 & 82.1 \\
        & RobustSAM         & 71.4 & 80.9 & \underline{78.1} & \underline{86.1} \\
        & \textbf{GaRA-SAM} & \textbf{74.8} & \textbf{83.4} & \textbf{80.0} & \textbf{87.4} \\

        \bottomrule
    \end{tabular}
    }
\vspace{-2mm}
\label{tab:bdd-lis}
\end{minipage}
\hfill
\hspace{0.5mm}
\begin{minipage}[t]{0.4\linewidth}
\centering
\captionsetup{font=footnotesize}
\caption{Evaluation results on BDD+LIS and ACDC, using a box prompt with ViT-L. GaRA-SAM-Syn and -Real are trained on Robust-Seg and BDD+LIS, respectively.}
\vspace{-2mm}
\renewcommand{\arraystretch}{0.9}
\resizebox{\linewidth}{!}{
\begin{tabular}{lcccc}
        \toprule
        \multirow{2}{*}{Method} 
        & \multicolumn{2}{c}{BDD+LIS} 
        & \multicolumn{2}{c}{ACDC} \\
        \cmidrule(lr){2-3} \cmidrule(lr){4-5}
        & IoU & Dice & IoU & Dice \\
        \midrule
        SAM               & 74.3 & 82.0 & 65.9 & 72.6 \\
        RobustSAM         & 78.1 & 86.1  & 67.5 & 77.0 \\
        \textbf{GaRA-SAM-Syn}  & \underline{80.0} & \underline{87.4} & \underline{71.6} & \underline{80.4} \\
        \textbf{GaRA-SAM-Real}  & \textbf{89.7} & \textbf{93.9} & \textbf{88.8} & \textbf{92.9} \\ 
        \bottomrule
    \end{tabular}
    }
\label{tab:realworld-finetune}
\vspace{-0.6mm}
\begin{center}
\caption{Comparison of our GaRA and LoRA on LVIS with ViT-B, using point prompts.}
\vspace{-2mm}
\renewcommand{\arraystretch}{0.9}
\resizebox{1.0\linewidth}{!}{
\begin{tabular}{llcccc}
        \toprule
        \multicolumn{2}{l}{\multirow{2}{*}{Method}} & \multicolumn{2}{c}{Degraded} & \multicolumn{2}{c}{Clear} \\
        \cmidrule(lr){3-4} \cmidrule(lr){5-6}
        & & IoU & Dice & IoU & Dice \\
        \midrule
        
        \multirow{3}{*}{LoRA}
        & Rank 16 & 75.0 & 84.3 & 79.4 & 87.3 \\
        & Rank 128 & \underline{75.6} &\underline{84.6} & \underline{80.6} & \underline{88.1} \\
        & Rank 256 & 73.4 & 83.1 & 78.5 & 86.6 \\
        \multicolumn{2}{c}{\textbf{GaRA-SAM}} & \textbf{77.0} & \textbf{85.7} & \textbf{81.3} & \textbf{88.7} \\
        \bottomrule
    \end{tabular}}\label{tab:lora-vs-gara}
\end{center}
\end{minipage}}
\vspace{-3mm}
\end{table*}

\subsection{Training on Real Corruption Datasets}
GaRA-SAM enables training on real corruption datasets without requiring access to clean references.
To demonstrate this, we train GaRA-SAM solely on the real-world dataset, BDD+LIS, and evaluate its performance on both the seen dataset (BDD+LIS) and an unseen real-world dataset, ACDC.
As shown in Table~\ref{tab:realworld-finetune}, training directly on real corrupted images significantly improves the performance on all benchmarks, outperforming the previous best by up to 21.3\% IoU on ACDC.
In addition, we compare GaRA-SAM trained on BDD+LIS (GaRA-SAM-Real) and that trained on the synthetic Robust-Seg dataset (GaRA-SAM-Syn), using box prompts and ViT-L in Table~\ref{tab:bdd-lis}; the results suggest the clear benefit of training directly on real corrupted images.

\begin{figure}[t!]
    \centering
    \includegraphics[width=\linewidth]{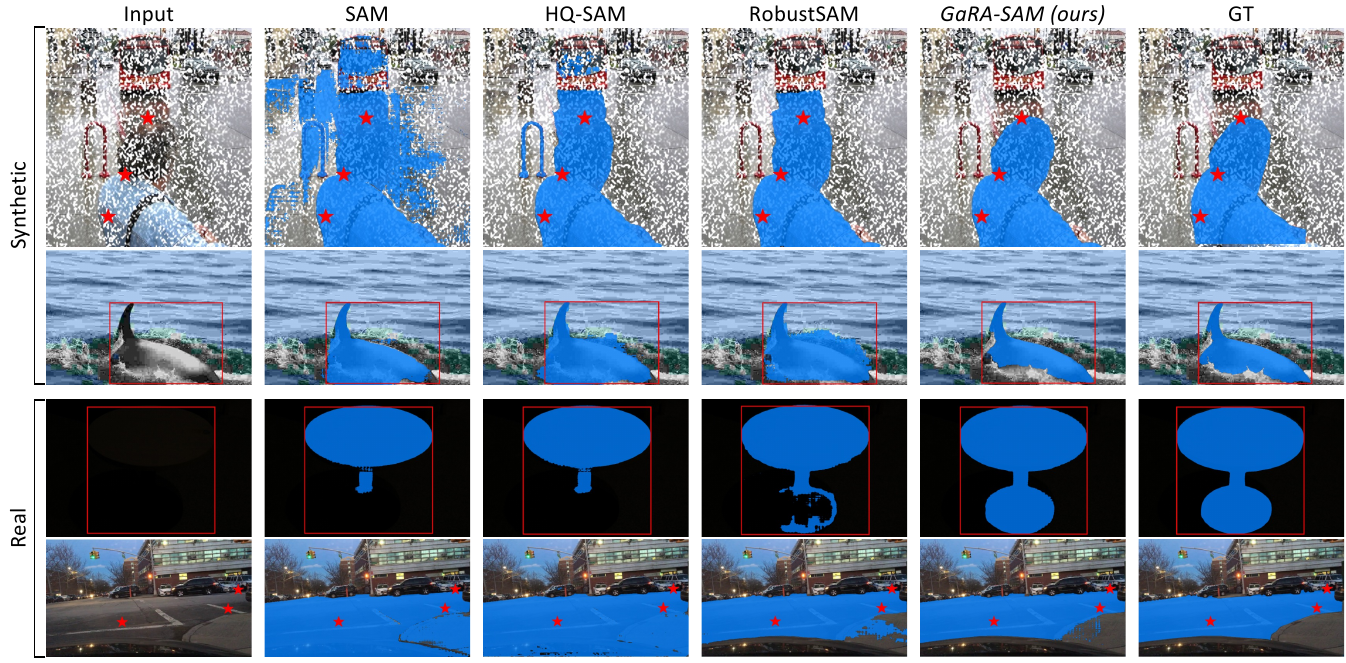}
    \vspace{-5mm}
    \caption{Results on synthetic (COCO, STREETS+NDD) and real corruption (BDD+LIS) datasets.
    }
    \label{fig:qual}
    \vspace{-2mm}
\end{figure}

\subsection{Qualitative results}
Figure~\ref{fig:qual} presents qualitative results under synthetic and real-world corruptions. GaRA-SAM delivers more accurate and complete masks than SAM, HQ-SAM, and RobustSAM, especially under severe degradations: it better preserves object boundaries in synthetic cases and is more reliable in low-light and adverse weather conditions. 

\subsection{In-depth analysis} \label{sec:in-depth-analysis}
We conduct a comprehensive ablation study to assess the contribution of each design choice in GaRA-SAM. All experiments are performed on the LVIS dataset, using point prompts for consistency.

\noindent\textbf{LoRA vs. GaRA.}
We first compare GaRA with the standard LoRA with fixed ranks.
As shown in Table~\ref{tab:lora-vs-gara}, GaRA-SAM consistently outperforms all fixed-rank LoRA. While LoRA establishes a strong baseline, its fixed-rank design limits flexibility. 
In contrast, GaRA-SAM dynamically composes (rank-1) components based on the input, resulting in improved
robustness.

\begin{table}[t]
\resizebox{\linewidth}{!}{
\begin{minipage}[t]{0.43\linewidth}
\centering
\captionsetup{font=footnotesize}
\caption{Comparison of GaRA and MoE LoRA. }
\renewcommand{\arraystretch}{1.06}
\resizebox{1.0\linewidth}{!}{
\huge
    \begin{tabular}{lcccc}
        \toprule
        \multirow{2}{*}{Method} & \multicolumn{2}{c}{Degraded} & \multicolumn{2}{c}{Clear} \\
        \cmidrule(lr){2-3} \cmidrule(lr){4-5}
        & IoU & Dice & IoU & Dice \\
        \midrule
        MoE LoRA (2, 16, 128, 256) & \underline{76.4} & \underline{85.3} & \underline{79.7} & \underline{87.5} \\
        MoE LoRA (2, 16, 128, 1024) & 75.4 & 84.5 & 78.6 & 86.7 \\
        \textbf{GaRA-SAM} & \textbf{77.0} & \textbf{85.7} & \textbf{81.3} & \textbf{88.7} \\
        \bottomrule
    \end{tabular}
    }
    \label{tab:abl_lora_moe}
\end{minipage}
\hfill
\hspace{0.5mm}
\begin{minipage}[t]{0.4\linewidth}
\centering
\captionsetup{font=footnotesize}
\caption{Effect of the rank space separation.}
\renewcommand{\arraystretch}{1.0}
\resizebox{\linewidth}{!}{
\huge
    \begin{tabular}{llcccc}
        \toprule
        \multicolumn{2}{l}{\multirow{2}{*}{Method}} & \multicolumn{2}{c}{Degraded} & \multicolumn{2}{c}{Clear} \\
        \cmidrule(lr){3-4} \cmidrule(lr){5-6}
        & & IoU & Dice & IoU & Dice \\
        \midrule
        \multirow{2}{*}{\shortstack[c]{w/o Gating 1}}
        & Rank 16 & \underline{76.2} & \underline{85.2} & \underline{79.9} & \underline{87.7} \\
        & Rank 256 & 73.3 & 82.8 & 77.3 & 85.5 \\   
        \midrule
        \multicolumn{2}{c}{w/ Gating 1} & \textbf{77.0} & \textbf{85.7} & \textbf{81.3} & \textbf{88.7} \\
        \bottomrule
    \end{tabular}
    }
    \label{tab:abl_rankspace}
\vspace{-2mm}
\end{minipage}}
\vspace{-4mm}
\end{table}

\noindent\textbf{MoE LoRA vs. GaRA.}
We also investigate an alternative approach to rank selection using a mixture-of-experts (MoE)~\cite{zhou2022mixture} variant of LoRA, where multiple fixed rank LoRA blocks are instantiated, and a shared gating module selects one for each input. 
As reported in Table~\ref{tab:abl_lora_moe}, both MoE variants perform reasonably well, but fall short of GaRA-SAM in both clean and degraded conditions.
This gap is not trivial regarding the fewer number of parameters of GaRA-SAM.

\noindent\textbf{Impact of Rank Space Selection.}
To evaluate the effectiveness of rank space selection (Gating 1), we compare GaRA-SAM with a variant that removes the separation between the lower- and higher-rank spaces, using a single unified set of (rank-1) components with a fixed maximum rank (\eg, 16 or 256). 
As shown in Table~\ref{tab:abl_rankspace}, eliminating rank space selection results in performance degradation, suggesting that separating the low- and high-rank components allows more effective specialization.

\noindent\textbf{Impact of Temperature in Gumbel-Sigmoid.}
We investigate the effect of the temperature of the Gumbel-Sigmoid by varying its value. The results in \Fig{ablation} suggest that GaRA-SAM remains stable across a wide range of temperature values.

\begin{table}[t]
    \centering
    \begin{minipage}{0.24\linewidth}
        \centering
        \includegraphics[width=\linewidth]{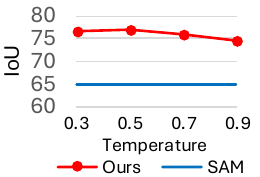} 
        \captionof{figure}{Effect of Gumbel temperature.}\label{fig:ablation}
    \end{minipage}
    \hspace{2mm}
    \begin{minipage}{0.7\linewidth}
        \centering
        \small
        \vspace{-3mm}
        \caption{Comparison of computational efficiency.}
        \begin{tabular}{lcccc}
            \toprule
            & \multicolumn{2}{c}{Training} & \multicolumn{2}{c}{Inference} \\
            \cmidrule(lr){2-3} \cmidrule(lr){4-5}
            Method & Learnable Params & \# GPU & GPU Memory & FPS \\
            \midrule
            SAM         & 1250M & 256 & 3.36GB & 2.9 \\
            RobustSAM   & 403M  & 8   & 5.41GB & 2.8 \\
            GaRA-SAM    & 343M  & 8   & 5.39GB & 2.6 \\
            \bottomrule
        \end{tabular}\label{tab:efficiency}
    \end{minipage}
    \vspace{-2mm}
\end{table}

\noindent\textbf{Computational Cost Analysis.}
We compare the computational efficiency of GaRA-SAM in terms of learnable parameters, GPU resources, memory consumption, and inference speed (FPS).
As summarized in Table~\ref{tab:efficiency}, GaRA-SAM achieves the lowest number of learnable parameters and GPUs, owing to its parameter-efficient design.
At inference time, it requires less GPU memory than RobustSAM, while maintaining comparable FPS to both SAM and RobustSAM.

\begin{figure}[t!]
    \vspace{-4mm}
    \centering
    \includegraphics[width=\linewidth]{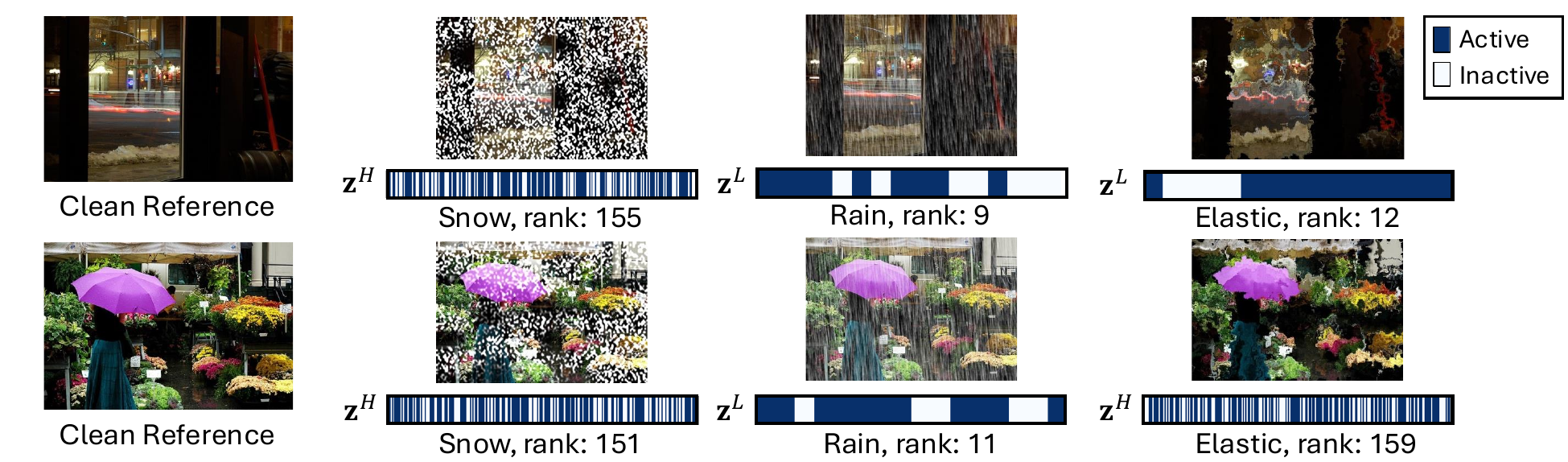}
    \vspace{-5mm}
    \caption{Visualization of the binary gating vectors and ranks under various images under corruptions.}\label{fig:analysis_garasam}
\vspace{-2mm}
\end{figure}
\vspace{-2mm}

\noindent\textbf{Analysis on Gating in GaRA-SAM.}
\Fig{analysis_garasam} demonstrates how the gating module responds to different combinations of corruption type and image. 
For each corrupted input, we present the corresponding binary activation vector $\mathbf{z}^H$ or $\mathbf{z}^L$ according to their selected rank space, along with the resulting number of active (rank-1) components.
We observe that GaRA activates different (rank-1) components depending on the image contents as well as the corruption types.
Even when the selected rank is similar, the constituent components often vary, highlighting the input-adaptive and fine-grained gating mechanism of GaRA.

\section{Conclusion} \label{sec:conclusion}
In this paper, we introduce GaRA-SAM, a novel approach for robustifying Segment Anything Model (SAM) under diverse image degradations. 
Through extensive empirical analysis, we observed that the optimal LoRA rank varies significantly across corruption types and individual inputs, motivating our design of Gated-Rank Adaptation (GaRA), a lightweight and input-adaptive module that dynamically modulates the effective rank of LoRA adapters. 
GaRA operates without requiring paired clean and degraded images, enabling fine-grained and parameter-efficient adaptation while preserving SAM’s inherent zero-shot generalization capabilities.
GaRA-SAM achieves state-of-the-art performance across all robustness benchmarks and, notably, supports training directly on real-world corruption datasets without clean references.
This leads to substantial gains in real-world scenarios, highlighting the practical utility and broad applicability of our method.


\clearpage

\bibliography{cvlab_kwak}
\bibliographystyle{plainnat}

\clearpage

\end{document}